\documentclass[sigconf,nonacm]{acmart}

\renewcommand\footnotetextcopyrightpermission[1]{} 

\AtBeginDocument{}

\begin{document}

\title{Towards an Action-Centric Ontology for Cooking Procedures Using Temporal Graphs}

\author{Aarush Kumbhakern}
\email{aarush.kumbhakern_ug25@ashoka.edu.in}
\orcid{0009-0003-4917-268X}
\authornote{These authors contributed equally.}
\authornote{Corresponding Authors}

\author{Saransh Kumar Gupta}
\email{saransh.gupta@ashoka.edu.in}
\orcid{0009-0000-5887-2301}
\authornotemark[1]
\authornotemark[2]

\author{Lipika Dey}
\email{lipika.dey@ashoka.edu.in}
\orcid{0000-0003-3831-5545}
\authornotemark[2]

\author{Partha Pratim Das}
\email{partha.das@ashoka.edu.in}
\orcid{0000-0003-1435-6051}

\affiliation{%
  \institution{Ashoka University}
  \city{Sonepat}
  \state{Haryana}
  \country{India}
}

\renewcommand{\shortauthors}{Kumbhakern et al.}

\begin{abstract}
  Formalizing cooking procedures remains a challenging task due to their inherent complexity and ambiguity. We introduce an extensible domain-specific language for representing recipes as directed action graphs, capturing processes, transfers, environments, concurrency, and compositional structure. Our approach enables precise, modular modeling of complex culinary workflows. Initial manual evaluation on a full English breakfast recipe demonstrates the DSL’s expressiveness and suitability for future automated recipe analysis and execution. This work represents initial steps towards an action-centric ontology for cooking, using temporal graphs to enable structured machine understanding, precise interpretation, and scalable automation of culinary processes - both in home kitchens and professional culinary settings.
\end{abstract}






\maketitle

\section{Introduction}

Recipes are ubiquitous procedural texts that describe multi-step transformations from raw ingredients to finished products. They interleave sequential and concurrent actions, environment changes, and domain-specific techniques, yet their apparent simplicity hides challenges for computation: ambiguous language, implicit context, and variable inputs/goals. Structured representations now underpin applications from automated cooking and robotic preparation to culinary knowledge bases and nutritional analysis. However, many existing formalisms lack the rigor, modular extensibility, or semantic precision needed for fine-grained reasoning about physical and temporal dynamics. Treating recipes as linear action lists often obscures state transitions, concurrency, spatial manipulation, and context-sensitive environmental effects.

We propose an early-stage framework that represents recipes as action graphs built from three atomic operations - \textsc{Process}, \textsc{Transfer}, and \textsc{Plate} - encoding cooking as a directed acyclic graph (DAG) of food-state and spatial transformations. Actions are parameterized by entities such as environment, tools, techniques, temperature profiles, and temporal constraints. Intermediate action outputs (partially processed components, PPCs) remain implicit, and feed directly into other actions without explicit extraction into nodes to keep graphs compact; full provenance is recoverable by traversing the local subgraph at each point of use. The framework natively models concurrency via branching with partial-order and relative-timing constraints, enabling interleaved actions (e.g. stirring while heating, adding an ingredient mid-process). Environments are explicit (and implicit, when associated with PPCs) context configurations supporting dynamic reassignment and spatial operations (e.g. tilting a pan to control spread/coverage).

The framework lays a foundation for simulation, robotic execution, automated instruction generation, and hybrid symbolic–neural reasoning. This paper details the design rationale, formal constructs, and representational choices of the action-graph grammar, outlines next steps toward large-scale parsing and execution, and offers a compact, extensible substrate for intelligent cooking systems, recipe understanding, and computational gastronomy.

\section{Related Work}

Early formalizations treated recipes as ordered imperative lists mapped to first-order action languages. MILK \cite{tasse2008sourcream}, introduced with the CURD corpus, encodes each instruction with a compact set of typed predicates (e.g. \textit{combine}, \textit{cut}, \textit{serve}) and tracks ingredient/tool state via creation/deletion. Despite showing (relatively) broad coverage, only $\approx 300$ recipes were annotated and automatic parsing accuracy was low, limiting downstream use.

Subsequent work sought richer structure from text. Flow Graphs \cite{mori2014flow} connect ingredients, tools, and actions with typed edges (200 annotated Japanese recipes) but conflate temporal and causal links and provide no released parser. Ingredient–Instruction Dependency Trees \cite{jermsurawong2015structure} attempt to extract the structure of recipes into an ``Ingredient Merging Map'', achieving 93\% accuracy on 260 English recipes; however, ingredient state, environment, and concurrency are out of scope. Large-scale information-extraction pipelines pair NER with dependency-based relation extraction to normalize ingredients and link techniques, utensils, and items \cite{diwan2020named}. Systems that induce reduced action tuples/graphs from instructions via NER + SRL likewise report persistent errors with coreference and with unstated or implicitly chosen tools/containers \cite{agarwal2011ie}.

To evaluate procedural competence, Nevens et al. \cite{nevens2024benchmark} map recipes to 38 executable kitchen actions and execute them in qualitative and quantitative simulators; their semantic language tracks environment state and parameters, but a single gold plan per recipe penalizes valid alternatives. Emerging latent program approaches jointly learn image/recipe/program embeddings \cite{papadopoulos2022programs} or mine sentence-level trees for generation/retrieval \cite{wang2023structural}, yet the induced programs remain opaque sequences lacking explicit environment transfers, collapsing intermediate states, and omitting concurrency.

Authoring-oriented DSLs emphasize metadata and validation. Corel \cite{roorda2021corel} types ingredients, tools, time, temperature, and nutrition and compiles to web pages, but omits environment tracking and time-relative processes.

Across these threads, gaps remain in (i) unifying state-altering and spatial transfers with environment lineage, (ii) representing partial products without graph bloat while retaining provenance, and (iii) expressing concurrent/interleaved actions with machine-checkable constraints. Our Recipe Action-Graph DSL addresses these: it separates \textsc{Process} (state change) from \textsc{Transfer} (spatial change) and binds every ingredient/PPC to an explicit environment lineage; keeps PPCs implicit yet fully traceable via backward traversal; and treats concurrency as first-class with merge semantics and relative-timing/interjection capabilities. A proposed technique lexicon with defaults and post-conditions ensures semantic consistency, and the graph supports multiple, equivalently valid execution plans.

\section{Methodology and Design}

The Recipe Action-Graph DSL encodes recipes as DAGs of parameterized action nodes, rather than as linear scripts. Nodes denote composable actions that transform inputs; edges capture both material flow and partial-order (relative-timing) constraints. The ontology that the DSL presupposes comprises three action types - \textsc{Process}, \textsc{Transfer}, and \textsc{Plate} - and the entity types Ingredients, Environments, and PPCs. Actions enact explicit state changes and environment-aware transfers, carrying raw inputs forward to \textsc{Plate}-ready outputs.

\paragraph{Graph compilation and semantics.} The DSL compiles to a DAG in which nodes represent actions and ingredients, while edges encode dependency relations. This structure permits multiple valid execution orderings (consistent with partial-order constraints) while preserving temporal precedence and maintaining ingredient provenance. Traceability is supported in both directions: from named raw ingredients to final dish components and from any point of use back to its sources.

\paragraph{Ingredients.} Ingredients follow a structured schema: name, quantity, unit, and optional form and modifiers. Only named ingredients serve as graph roots; all subsequent transformations flow from these initial entities.

\paragraph{Environments.} Every item in the system (ingredient or PPC) may be associated with an \emph{environment}, modeled as a tuple - \texttt{(container, location, optional geometry)} - where geometry captures pose or orientation (e.g. tilt angle). This makes explicit not only \emph{what} is acted upon, but also \emph{where} and \emph{how} it is situated - a critical determinant of culinary outcomes.

\begin{figure}[h]
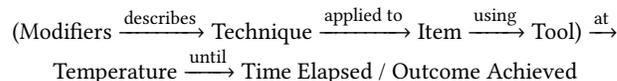

    \centering
    (Modifiers $\xrightarrow{\text{describes}}$ Technique $\xrightarrow{\text{applied to}}$ Item $\xrightarrow{\text{using}}$ Tool) $\xrightarrow{\text{at}}$ Temperature $\xrightarrow{\text{until}}$ Time Elapsed / Outcome Achieved
    \caption{Generalized high-level characterization of cooking processes.}
\end{figure}

\paragraph{Action types} 

\begin{itemize}
    \item \textsc{Process} nodes encode changes to physical/chemical state, parameterized by an input (either an Ingredient or a PPC), technique, tool, temperature specification (with ramp/curve support), time/duration, completion conditions, and modifiers. The technique parameter is a reference to an object entry in a formalized culinary technique lexicon, containing a standardized static and optional parametrized definition (see Future Work). Each process step can include rich sensory and outcome-based termination conditions - such as doneness, texture, color, or aroma - allowing for flexible, yet grounded reasoning about process completion.

    \item \textsc{Transfer} nodes encode source $\to$ destination environment reassignment of an Ingredient or PPC between environments (e.g. to a pan, oven, or fridge). Each transfer updates the subsequent PPC's associated environment for all subsequent actions, modeling context-driven state changes (implying that a \textsc{Transfer} can then occur from one PPC to another, with the PPC produced by the transfer inheriting the environment of the target PPC).

    \item \textsc{Plate} nodes are reserved for final touches - pure assembly and presentation (currently undeveloped).
\end{itemize}

\paragraph{Intermediate state and provenance} Intermediate states are implicit. PPCs are the unnamed outputs of \textsc{Process} and \textsc{Transfer} nodes and are consumed directly by subsequent actions. By omitting explicit intermediate item nodes, the design remains compact while preserving full information and traceability: the provenance of any component can be reconstructed via backward traversal of the subgraph rooted at its point of use.

\paragraph{Global axioms and validation} Global invariants include: (i) acyclicity of the top-level action graph; (ii) type safety for all action parameters; and (iii) persistence of an item's environment association until an explicit re-\textsc{Transfer}. Technique references are validated against a controlled lexicon with defaults and post-conditions. A recipe always yields a single final output but may aggregate multiple PPCs (e.g. a main, sauce, and garnish) in its final assembly.

\paragraph{Concurrency and relative timing} Concurrency is represented natively by parallel branches that synchronize at explicit merge points. The simplest form of concurrency is the implicit choice of concurrent execution of multiple mutually independent components of a recipe. For example, a recipe may state ``while the cake is baking, we can get started on the buttercream''. This presents two completely separate execution branches - making a cake, and making buttercream - that eventually merge at the point of icing, in this case. Concurrent \emph{actions} on the other hand, such as interleaved steps and time-relative interjections (e.g. ``add garlic halfway through sautéing'') are modeled via relative-timing fields and sub-processes, providing expressive control over multitasking while respecting the DAG's partial order. Interjection behaviors are encoded inside node semantics that determine, for example, whether the interjection is a breaking process (pauses the parent process), or repeating (e.g. ``add cold water every two minutes'').   

\paragraph{Extensibility and modularity} Actions, and environments reference external, versioned object definitions (e.g. techniques), allowing new elements to be introduced without altering the core schema. Recipes themselves can be embedded as \emph{plugin nodes} within other recipes, supporting modular construction of complex dishes; for example, a validated ``poached egg'' recipe can be inserted as a reusable node in a larger breakfast dish.

\section{Conceptual and Feasibility Illustration}

To qualitatively assess the expressiveness and practical utility of the Recipe Action-Graph DSL, we performed a manual encoding of a canonical, multi-component dish: the full English breakfast. We selected a detailed recipe from a food blog \cite{le2019englishbreakfast} for its rich procedural structure, heterogeneous tools and environments, multiple concurrent pipelines, and frequent reassignments of PPCs. This setting provides a demanding test of whether the early-stage DSL can faithfully represent real-world cooking logic rather than merely stepwise narration. Our effort focuses on how each aspect of this dish is captured in the action graph, demonstrating the DSL’s ability to encode complex cooking workflows, synchronize parallel activities, and preserve all information present in the original instructions. Below, we highlight key aspects of the recipe and describe the corresponding constructs in our action graph formalism.

\begin{figure}[h]
  \centering
  \includegraphics[width=\linewidth]{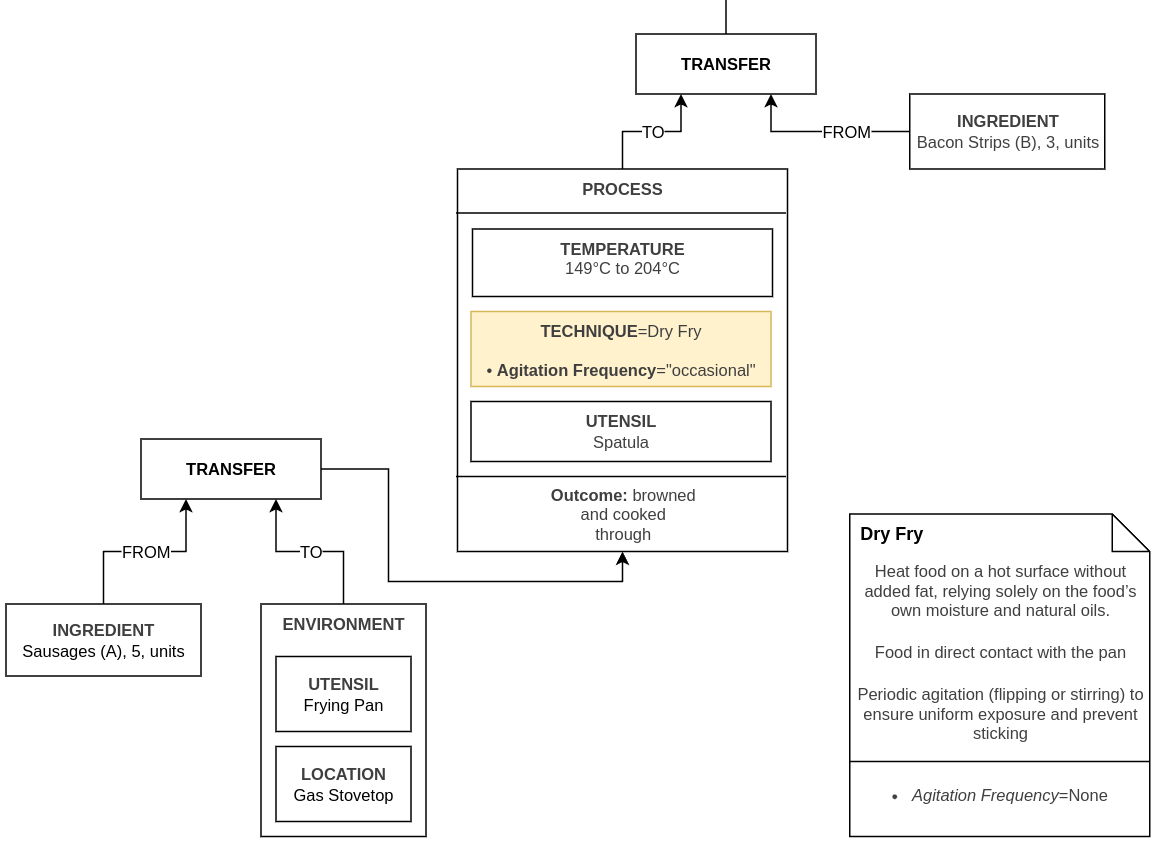}
  \caption{Visual excerpt describing cooking sausages from full English breakfast recipe action graph}
  \label{fig:exemplar}
\end{figure}

In Figure \ref{fig:exemplar}, the environment node encodes a \emph{(container, location)} tuple. The ingredient node ``Sausages [A]'' illustrates per-item identity: lettered instance tags disambiguate multiple items sharing a container (e.g. sausages [A] and bacon [B]) so they may later be referenced and processed independently. Directed edges labeled ``from'' (ingredient) and ``to'' (environment) converge on a \textsc{Transfer} node that records the placement of the sausages from their initial environment (unassigned/$\phi$) into the pan on the stove. Elevating this placement to an explicit \textsc{Transfer} makes the material flow - often implicit in textual recipes - concrete and updates the item’s associated environment for all subsequent actions. The following stage is a \textsc{Process} node, where the actual cooking of the sausages takes place. Key features encoded here include:

\begin{itemize}
    \item \textbf{Temperature Block:} The inclusion of a numeric Celsius range (inferred from ``medium heat'' + ``fry'') reinforces the DSL's design to translate subjective culinary terminology into precise, machine-interpretable parameters - aided by automated or LLM-supported normalization of temperature terms during parsing.

    \item \textbf{Technique Reference:} Pointing to the formal lexicon entry \verb|dry_fry| gives unambiguous semantics and reusable defaults. Technique-specific parameters \\ (e.g. \verb|agitation_frequency=occasional|) add controlled granularity while keeping the step tied to a single, well-defined operation with documented pre/post-conditions.

    \item \textbf{Utensil Block:} Declaring a spatula as the manipulating tool captures \emph{how} the item is acted upon, not just \emph{what} is cooked. Tool identity can constrain feasible techniques, influence geometry (turning/flip actions), and support later targets like robotic execution or capacity checks.

    \item \textbf{Termination Condition:} An outcome-type predicate (``browned and cooked through'') encodes observational endpoints alongside (or instead of) fixed durations. These can map to sensor-detectable signals or human-observable criteria, allowing flexible yet testable completion logic.
\end{itemize}

\begin{figure}[h]
  \centering
  \includegraphics[width=\linewidth]{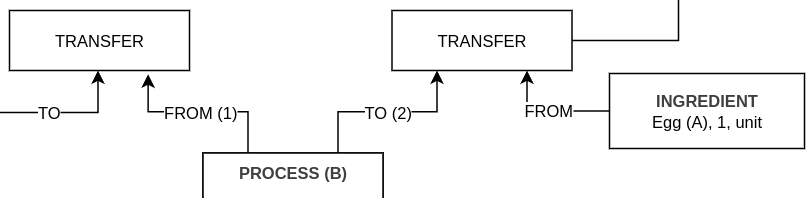}
  \caption{Visual excerpt describing sequential branching from full English breakfast recipe action graph}
  \label{fig:sequence}
\end{figure}

In the English-breakfast action graph, Figure \ref{fig:sequence} shows a process node whose output fans out to two successive \textsc{Transfer}s. The first outgoing edge (labeled \textsc{from (1)}) removes the just-processed PPC from its current environment (e.g. evacuating a cooked item from the pan). The second outgoing edge (labeled \textsc{to (2)}) introduces a different ingredient into that very environment, reusing the freed resource. This ordered \textsc{from} $\to$ \textsc{to} pattern makes environment occupancy and sequencing explicit in the DAG: it encodes that the pan must be vacated before the next item enters, enabling precise scheduling, resource contention analysis, and faithful reproduction of staging maneuvers like decanting or reserving fat before adding new ingredients. Typical cases include draining off rendered fat or fond prior to introducing tomatoes or mushrooms, or sliding toast into the pan immediately after the bacon exits - interdependent cycles through a shared container that are often only implied in prose recipes.

We also note current limitations in \textsc{Transfer} semantics. It is not yet formalized whether a transfer targets: (i) a specific PPC instance; (ii) an aggregated set of components; (iii) `all residuals' remaining in an environment (e.g. fat, fond); or (iv) a purely logical state handoff with no material flow. Likewise, parameters for quantity and granularity (whole vs. partial transfer; absolute units vs. proportions), and for transfer mode (manual removal, pouring, scraping, ladling, draining, tongs/lift) are underspecified. These choices have downstream effects - on provenance, on environment state (what residue remains), and on scheduling (atomicity/locking of a shared container). In future iterations, we plan to make these dimensions explicit.

\begin{figure}[h]
  \centering
  \includegraphics[width=\linewidth]{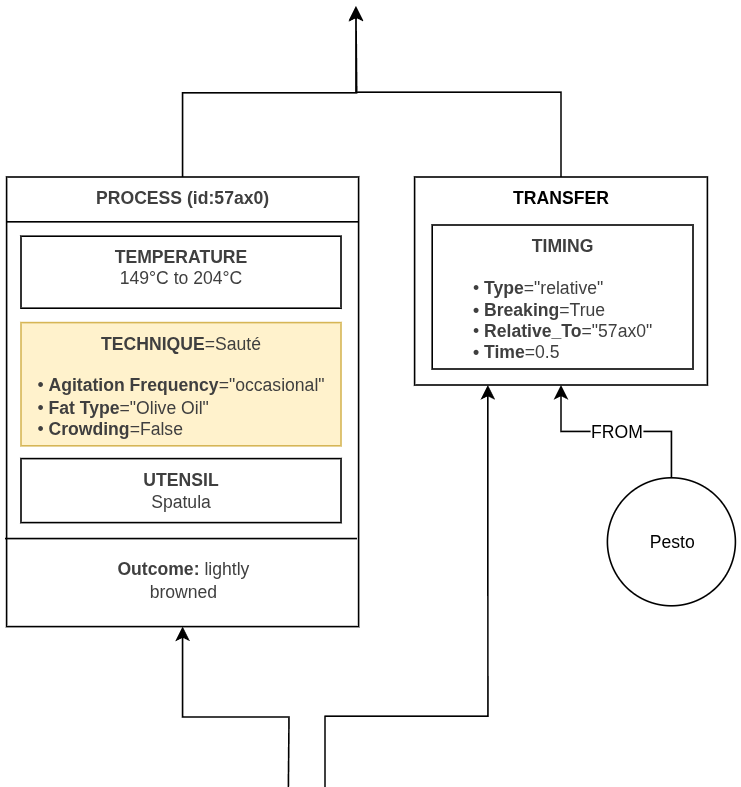}
  \caption{Concurrent process action graph snippet}
  \label{fig:concurrent}
\end{figure}

Figure \ref{fig:concurrent} deigns to roughly exemplify how a concurrent, relatively timed process would be represented, with the example of adding Pesto midway through lightly sautéing vegetables. In addition, we demonstrate here an aspect of the extensibility architecture, by plugging in Pesto as a final product node (hence circular), which would expand if necessary to a pre-existing Pesto recipe.

\section{Comparative Analysis}

We adopt a hybrid quantitative–qualitative evaluation to test whether the proposed Action-Graph DSL captures \emph{procedural} cooking semantics more faithfully than representative baselines with distinct aims. We compare against three public formalisms: (i) \textbf{MILK} \cite{tasse2008sourcream} - a compact, semantically oriented predicate language; (ii) \textbf{Corel} \cite{roorda2021corel} - a modern, author-first DSL emphasizing typed ingredients, time and temperature annotations, and validation; and (iii) \textbf{Culinary Grammar (Bagler)} \cite{bagler2022generative} - a generative phrase-structure specification built from ``cubits'' (culinary concepts) with rule-based blending. A purely quantitative tally risks rewarding authoring conveniences over execution semantics; a purely qualitative read risks cherry-picking. We therefore combine (i) a feature-by-feature coverage rubric and (ii) targeted behavioral probes stressing concurrency, environment lineage, interjections/relative timing, and observational termination. Prior characterizations motivate this split: Corel offers explicit time/temperature syntax and checking but omits environments, transfers, and concurrency; MILK maintains a minimal action set and a container/world model but encodes time/temperature as strings and lacks concurrency; Culinary Grammar (Bagler) formalizes ingredient/quantity/form/process/descriptor/utensil and recursive composition but provides no explicit environment lineage, concurrency, or transfer semantics.

\subsection{Quantitative rubric and Qualitative Protocol}

We score each DSL on a 2/1/0 scale - \textbf{2} (explicit/first-class), \textbf{1} (implicit/partial), \textbf{0} (unsupported) - across a 29-item checklist spanning entities/metadata, process/actions, temporal/thermal, graph/control, and validation/tooling. Totals are expressed as both raw points and coverage percentage; the denominator (58) reflects the number of features evaluated at a 2-point maximum each.

We instantiate realistic behaviors and ask what each formalism can \emph{natively} express: keeping items warm via environmental state; pan reuse and staging via \textsc{Transfer} with residue handling; outcome-based termination (e.g. ``browned and cooked through''); and multitasking with interjections (``add halfway''). We judge fidelity without falling back to untyped prose.

\begin{table*}[t]
    \centering
    \setlength{\tabcolsep}{8pt}

    \caption{Coverage summary using the 2/1/0 rubric (higher \% = broader first-class coverage).}
    \label{tab:coverage-summary}
    
    \begin{tabular*}{\textwidth}{@{\extracolsep{\fill}}lrrrrr}
        \toprule
        \textbf{DSL / Grammar} & \textbf{Explicit (2)} & \textbf{Implicit (1)} & \textbf{Missing (0)} & \textbf{Score (/58)} & \textbf{Coverage} \\
        \midrule
        MILK \cite{tasse2008sourcream}                & 9  & 9  & 11 & \textbf{27} & \textbf{46.6\%} \\
        Corel \cite{roorda2021corel}                  & 7  & 4  & 18 & \textbf{18} & \textbf{31.0\%} \\
        Culinary Grammar (Bagler) \cite{bagler2022generative} & 9  & 7  & 13 & \textbf{25} & \textbf{43.1\%} \\
        Action-Graph (ours)                           & 19 & 4  &  6 & \textbf{42} & \textbf{72.4\%} \\
        \bottomrule
    \end{tabular*}

    \medskip
    \emph{Notes.} MILK emphasizes a compact predicate set and container/world relations; Corel emphasizes authoring and numerics; Bagler’s grammar formalizes lexical and phrase-structure regularities (cubits; IP/PP/S recursion) with a large corpus; the Action-Graph DSL targets procedural fidelity (explicit \textsc{Process}/\textsc{Transfer}/\textsc{Plate}, environments, concurrency, relative timing, temperature ramps, implicit-but-recoverable PPCs).
\end{table*}

\subsection{Results}

\paragraph{Coverage.} Quantitatively, Action-Graph attains the highest breadth of coverage (42/58, 72.4\%), followed by MILK (27/58, 46.6\%), \emph{Culinary Grammar (Bagler)} (25/58, 43.1\%), and Corel (18/58, 31.0\%). Bagler’s score is driven by explicit entities and a typed process lexicon, plus descriptors that capture outcomes and time expressions; it lacks explicit modeling of environment lineage, transfer semantics, concurrency, and provenance.

\paragraph{Behavioral expressiveness.} Corel offers native numeric time and temperature but no relative timing; MILK uses strings/predicates; Culinary Grammar (Bagler) encodes durations and qualitative descriptors (e.g. ``for 10 minutes'', ``preheat 220°C''), but without typed ramps or relative offsets; Action-Graph adds dedicated fields and ramps and supports interjections. Neither MILK, Corel, nor Bagler includes a concurrency construct; Action-Graph provides first-class concurrency with merges. MILK has container/location relations but not full lineage; Corel leaves environments in prose; Bagler models utensils and ``in X'' prepositional attachment but no explicit environment modeling or transfer updates; Action-Graph binds and updates an item’s environment on \textsc{Transfer}, enabling staging and traceable provenance.

\paragraph{Where Action-Graph excels.} Our DSL remains the only execution-representation DSL, and uniquely operationalizes execution semantics (state vs.\ spatial change, environment lineage, partial orders with interjections, temperature profiles, and concurrency) while retaining compactness via implicit PPCs.

\section{Future Work}

The Recipe Action-Graph DSL is the terminal representation in a three-stage pipeline that aims to convert natural-language recipes into structured, machine-interpretable, and action-centric ontologies and temporal graphs. The DSL specification defines the target schema; the end-to-end implementation couples large language models with classical NLP to populate it.

\begin{enumerate}
    \item \textbf{Simplification:} An LLM performs information-preserving rewriting to produce minimal, step-by-step instructions: it removes redundancies, clarifies ambiguous references, and splits compound clauses into sequential atomic steps - without dropping semantic content.

    \item \textbf{Standardization:} A second LLM normalizes entities against two comprehensive culinary lexicons. Ingredient names are mapped to canonical labels; techniques are mapped to standardized entries with parameter slots. For example, ``pour fat or juices over (meat) during cooking to keep it moist'' is normalized to the technique baste, with parameters for agent, target, and frequency.

    \item \textbf{Parsing:} Domain-adapted NER and structured IE then instantiate the action graph: ingredients, techniques, tools, and temporal markers are detected; \textsc{Process}/\textsc{Transfer}/\textsc{Plate} nodes are created with complete parameters, including technique references, environment context, temperature profiles (with ramps/curves where applicable), and temporal constraints (absolute and relative).
\end{enumerate}

In the processing pipeline, an Implicit Step Recovery module is integrated after recipe simplification and standardization but before final action graph construction. This module is designed to identify and reconstruct essential procedural steps that are not explicitly stated in the recipe text but are required for correct execution and complete formalization. 

\section{Conclusion}

We presented a modular, extensible DSL that models recipes as structured action graphs built from three atomic nodes - \textsc{Process}, \textsc{Transfer}, and \textsc{Plate}. The design supports implicit state tracking (PPCs), explicit environment modeling, and first-class concurrency, providing a robust foundation for capturing the procedural and contextual richness of real-world instructions.

Looking ahead, we will pair large language models with classical NLP to normalize and simplify recipe text, standardize entities, recover implicit steps, and compile graphs. This pipeline bridges informal, variable prose to machine-interpretable structure, enabling culinary knowledge analysis, automated execution, and digital authoring.

Future work includes large-scale extraction, construction of a technique lexicon (with defaults and postconditions), and systematic validation of expressiveness and utility across diverse recipe types. Advancing this formal modeling aims to catalyze research in computational gastronomy and intelligent kitchen technologies.

\begin{acks}
    This research was supported by the Mphasis AI \& Applied Tech Lab at Ashoka - a collaboration between Ashoka University and Mphasis Limited.
\end{acks}

\bibliographystyle{ACM-Reference-Format}
\bibliography{main}

\end{document}